\documentclass[10pt,twocolumn,letterpaper]{article}

\usepackage[pagenumbers]{cvpr} % To force page numbers, e.g. for an arXiv version

\usepackage{graphicx}
\usepackage{amsmath}
\usepackage{amssymb}
\usepackage{booktabs}
\usepackage[dvipsnames]{xcolor}
\usepackage{flexisym}
\usepackage{amsmath}
\usepackage{multirow}
\usepackage{tabularx}
\usepackage{enumitem}
\setenumerate[1]{itemsep=0pt,partopsep=0pt,parsep=\parskip,topsep=5pt}
\setitemize[1]{itemsep=0pt,partopsep=0pt,parsep=\parskip,topsep=5pt}
\setdescription{itemsep=0pt,partopsep=0pt,parsep=\parskip,topsep=5pt}

\usepackage{amssymb}% http://ctan.org/pkg/amssymb
\usepackage{pifont}% http://ctan.org/pkg/pifont
\usepackage[pagebackref,breaklinks,colorlinks]{hyperref}

\usepackage[capitalize]{cleveref}
\crefname{section}{Sec.}{Secs.}
\Crefname{section}{Section}{Sections}
\Crefname{table}{Table}{Tables}
\crefname{table}{Tab.}{Tabs.}

\def\cvprPaperID{6593} % *** Enter the CVPR Paper ID here
\def\confName{CVPR}
\def\confYear{2022}

\begin{document}

\title{MonoPLFlowNet: Permutohedral Lattice FlowNet for Real-Scale 3D Scene Flow Estimation with Monocular Images}
\author{Runfa Li, Truong Nguyen\\
Department of Electrical and Computer Engineering, University of California, San Diego\\
{\tt\small \{rul002, tqn001\}@eng.ucsd.edu}
\and
}
\twocolumn[{
\maketitle
\vspace{-10mm}
\begin{center}
    \captionsetup{type=figure}
    \begin{minipage}{0.29\linewidth}
  \centerline{\includegraphics[width=1.0\textwidth]{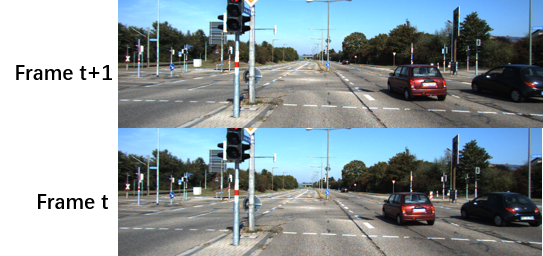}}
\end{minipage}
\hfill
\begin{minipage}{0.39\linewidth}
  \centerline{\includegraphics[width=1.0\textwidth]{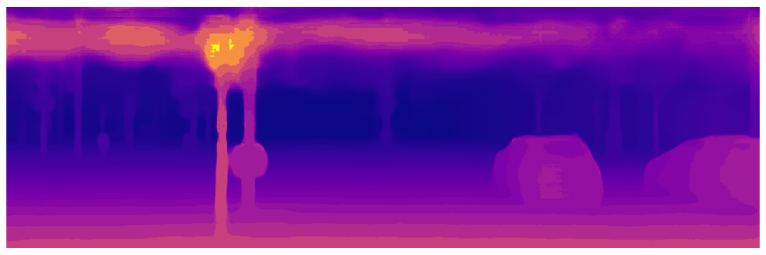}}
\end{minipage}
\hfill
\begin{minipage}{0.31\linewidth}
  \centerline{\includegraphics[width=1.0\textwidth]{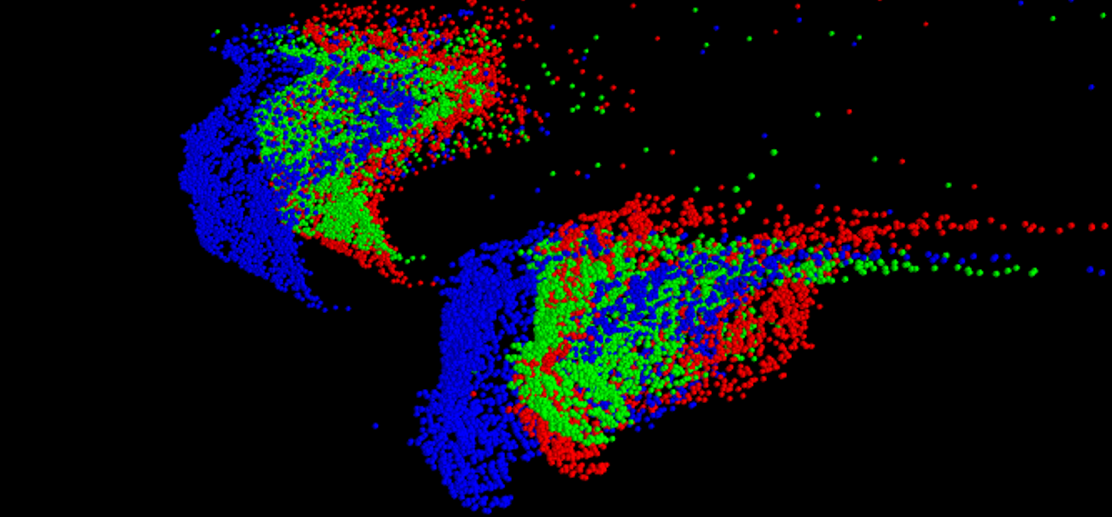}}
\end{minipage}
\vspace{-3mm}
    \captionof{figure}{\textbf{Results of our MonoPLFlowNet.} With only two consecutive monocular images (\textit{left}) as input, our MonoPLFlowNet  estimates both the depth (\textit{middle}) and 3D scene flow (\textit{right}) in real scale. \textit {Right} shows a zoom-in real-scale scene flow of the two vehicles from side view with the pseudo point cloud generating from the estimated depth map (\textit(middle), where blue points are from frame \textit{t}, red and green points are blue points translated to frame \textit{t}+1 by ground truth and estimated 3D scene flow, respectively. The objective is to align green and red points.}
\end{center}
}]

\maketitle
%%%%%%%%% ABSTRACT
\begin{abstract}
Real-scale scene flow estimation has become increasingly important for 3D computer vision. Some works successfully estimate real-scale 3D scene flow with LiDAR.  However, these ubiquitous and expensive sensors are still unlikely to be equipped widely for real application. Other works use monocular images to estimate scene flow, but their scene flow estimations are normalized with scale ambiguity, where additional depth or point cloud ground truth are required to recover the real scale. Even though they perform well in 2D, these works do not provide accurate and reliable 3D estimates. We present a deep learning architecture on permutohedral lattice - MonoPLFlowNet. Different from all previous works, our MonoPLFlowNet is the first work where only two consecutive monocular images are used as input, while both depth and 3D scene flow are estimated in real scale. Our real-scale scene flow estimation outperforms all state-of-the-art monocular-image based works recovered to real scale by ground truth, and is comparable to LiDAR approaches. As a by-product, our real-scale depth estimation also outperforms other state-of-the-art works. Code will be available at \url{https://github.com/BlarkLee/MonoPLFlowNet}.
\end{abstract}

\begin{figure*}[t]
    \centering
    \includegraphics[width=\linewidth]{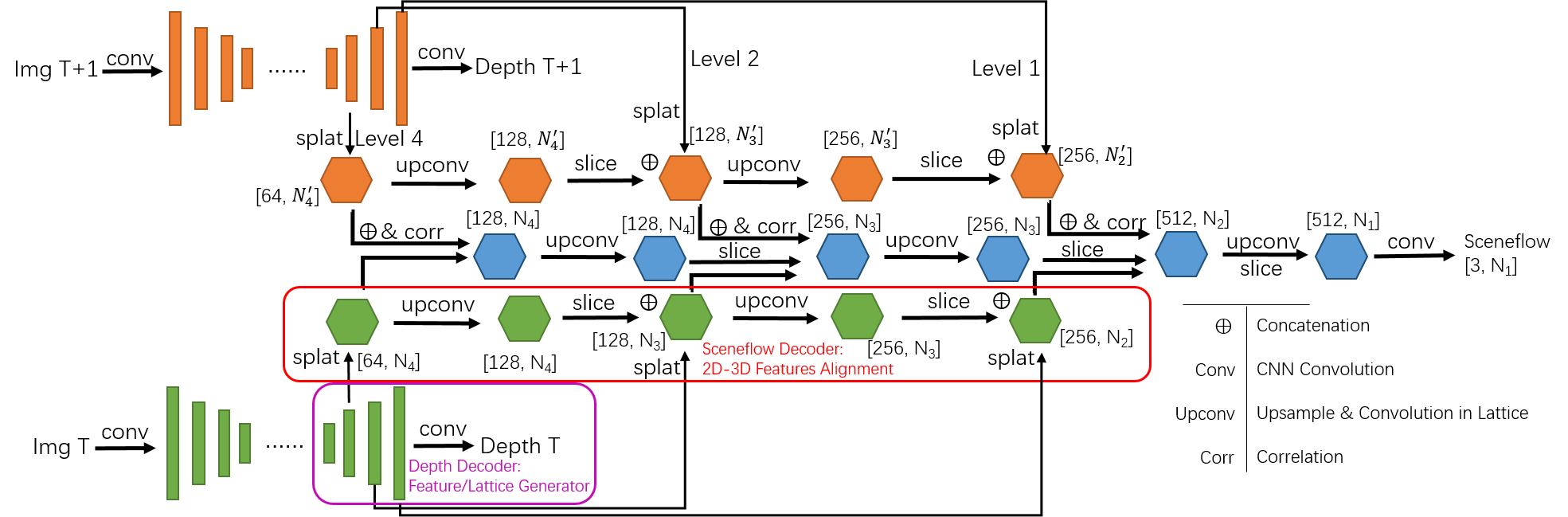}
    \vspace{-7mm}
    \caption{\textbf{MonoPLFlowNet Architecture: An Hour-Glass like encoder-decoder based network.}  It shares the same encoder for two monocular images as the only input, and jointly estimates the depth and 3D scene flow in real scale by decoding separately with a {\color{Purple}\textit{Depth decoder (purple box)}} and a {\color{red}\textit{Sceneflow Decoder (red box)}}. Architectures of the two decoders are shown in Figure \ref{fig:depth_decoder} and \ref{fig:sceneflow_decoder}, respectively.} 
    \label{fig:MonoPLFlowNet}
    \vspace{-5mm}
\end{figure*}

%%%%%%%%% BODY TEXT
\vspace{-5mm}
\section{Introduction}
\label{sec:intro}

Scene flow are 3D vectors associating the corresponding 3D point-wise motion between consecutive frames, where scene flow can be recognized as lifting up pixel-wise 2D optical flow from the image plane to 3D space. Different to coarsely high-level motion cues such as bounding-box based tracking, 3D scene flow focuses on precisely low-level point-wise motion cues. With such advantage, scene flow can either serve for non-rigid motion as visual odometry and ego motion estimation, or rigid motion as multi-object tracking, which makes it increasingly important in motion perception/segmentation and applications in dynamic environments such as robotics, autonomous driving and human-computer interaction. 

3D scene flow has been widely studied using LiDAR point cloud  \cite{flownet3d, flownet3d++, splatnet, pointflownet, hplflownet, pointpwc, justgowithflow} from two consecutive frames as input, where a few recent LiDAR works achieve very accurate performances. However, LiDARs are still too expensive to be equipped for real applications. Other sensors are also being explored for 3D scene flow estimation such as RGB-D cameras \cite{learningrigidity, sphereflow, motioncoorperation} and stereo cameras\cite{stereomulti, sceneflowfields, howimportant, sense, instanceflow}. However, each sensor configuration also has its own limitation, such as RGB-D cameras are only reliable in the indoor environment, while stereo cameras require calibration for stereo rigs.

Since Monocular camera is ubiquitous and cheap for all real applications, it is a promising alternative to the complicated and expensive sensors. There are many works for monocular image-based scene flow estimation\cite{geonet,dfnet,CC,GLNet,epc,epc++, mono_expansion, monosf, monosf_multi}, where CNN models are designed to jointly estimate monocular depth and optical/scene flow. However, their estimations are all with scale ambiguity. This problem exists in all monocular works where they estimate normalized depth and optical/scene flow. To recover to the real scale, they require depth and scene flow ground truth, which are possible to obtain for evaluation in labeled datasets but impossible for a real application.

Our motivation for this work is to take the advantages and overcome the limitations from both LiDAR-based (real-scale, accurate but expensive) and image-based (cheap, ubiquitous but scale-ambiguous) approaches. Our key contributions are:
\begin{itemize}[leftmargin=0cm]
\setlength{\itemsep}{0pt}
\setlength{\parsep}{0pt}
\setlength{\parskip}{0pt}
\item We build a deep learning architecture MonoPLFlowNet, which is the first work using only two consecutive monocular images to estimate in real scale both the depth and 3D scene flow. 
\item Our 3D scene flow estimation outperforms all state-of-the-art monocular image-based works even after recovering their scale ambiguities with ground truth, and is comparable to LiDAR-based approaches.
\item  We introduce a novel method - "Pyramid-Level 2D-3D features alignment between Cartesian Coordinate and Permutohedral Lattice Network", which bridges the gap between features in monocular images and 3D points in our application, and inspires a new way to align 2D-3D information for computer vision tasks which could benefit other applications.
\item As a byproduct, our linear-additive depth estimation from MonoPLFlowNet also outperforms state-of-the-art works on monocular depth estimation.
\end{itemize}

%-------------------------------------------------------------------------

%------------------------------------------------------------------------
\section{Related Works}
\noindent\textbf{Monocular image-based scene flow estimation:} Monocular 3D scene flow estimation originates from 2D optical flow estimation\cite{flownet-c, mono_expansion}. To get real-scale 3D scene flow from 2D optical flow, real-scale 3D coordinates are required which could be derived from real-scale depth map. Recently, following the success from SFM (Structure from Motion) \cite{sfmlearner, sm3d}, many works jointly estimate monocular depth and 2D optical flow \cite{geonet, dfnet, CC, GLNet, epc, epc++}. However, as seen in most SFM models, the real scale decays in jointly training and leads to scale ambiguity. Although \cite{monosf, monosf_multi} jointly estimate depth and 3D scene flow directly, they suffer from scale ambiguity, which is the biggest issue for monocular image-based approaches.

\vspace{1mm}
\noindent\textbf{3D Point cloud based scene flow estimation:} Following PointNet\cite{pointnet} and PointNet++\cite{pointnet++}, it became possible to use CNN-based models to directly process point cloud for different tasks including 3D scene flow estimation. Since directly implementing on 3D points, there is no scale ambiguity. The success of CNN-based 2D optical flow networks also boost 3D scene flow.  FlowNet3D\cite{flownet3d} builds on PointNet++ and imitates the process used in 2D optical FlowNet\cite{flownet-c} to build a 3D correlation layer. PointPWC-net\cite{pointpwc} imitates another 2D optical flow network PWC-net\cite{pwcnet} to build 3D cost volume layers. The most successful works are "Permutohedral Lattice family", where BCL\cite{BCL}, SplatNet\cite{splatnet}, HPLFlowNet\cite{hplflownet}, PointFlowNet\cite{pointflownet} all belong to the family. The lattice (more details will be provided in Section 3.1) is a very efficient representation for processing of high-dimensional data \cite{fast_filtering}, including 3D point cloud. Point cloud based methods achieves better performance than image-based methods. However, with LiDAR scanning as the input, they are still too expensive for real applications.

Our MonoPLFlowNet takes advantages and overcome limitations from both image (cheap, ubiquitous but scale-ambiguous) and LiDAR (real-scale, accurate but expensive) based approaches by only using monocular images to jointly estimate depth and 3D scene flow in real scale.

%------------------------------------------------------------------------
\section{MonoPLFlowNet}

\begin{figure*}[t]
    \centering
    \includegraphics[width=\linewidth]{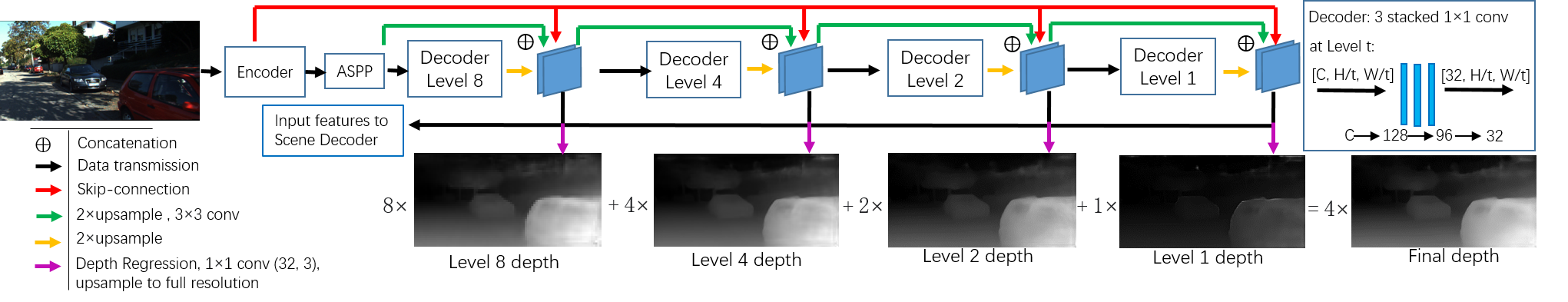}
    \vspace{-7mm}
    \caption{\textbf{Depth Decoder: Pyramid-Level Linearly-Additive Depth Estimation.} Besides the accurate depth estimation, from a whole architecture view, our depth decoder also serves as the feature/lattice position generator, which is why we design it in linear-additive way.} 
    \label{fig:depth_decoder}
    \vspace{-5mm}
\end{figure*}

Our MonoPLFlowNet is an hour-glass encoder-decoder based model, which only takes two consecutive monocular images as input, while both the depth and 3D scene flow are estimated in real scale. Figure \ref{fig:MonoPLFlowNet} shows our network architecture. While sharing the same encoder, we use separate decoders for depth and scene flow. With our designed mechanism, we align the 2D-3D features to boost the performance of each other. In this section, we present the theory of permutohedral lattice, and then discuss how we design and align the two decoders.

\subsection{Review of Permutohedral Lattice Filtering}
\noindent
\textbf{High-Dimensional Gaussian Filtering:}
Signal's value and position are two important  aspects of filtering. Eq. \ref{eq:gaussian} shows a general form of high-dimensional Gaussian filtering:
\vspace{-2mm}
\begin{equation}
    \vec{v_i} = \sum_{j=1}^{n} exp^{-\frac{1}{2}(\vec{p_i}-\vec{p_j})^T\sum^{-1}(\vec{p_i}-\vec{p_j})^T } \vec{v_j}
\label{eq:gaussian}
\end{equation}

\vspace{-5mm}
\begin{equation}
    \vec{v_i} = (r_i,g_i,b_i, 1),  \;
    \vec{p_i} = (\frac{x_i}{\sigma_s}, \frac{y_i}{\sigma_s})
    \label{eq:gaussian_blur}
\end{equation}

\vspace{-3mm}
\begin{equation}
    \vec{v_i} = (r_i,g_i,b_i, 1),  \;
    \vec{p_i} = (\frac{x_i}{\sigma_s}, \frac{y_i}{\sigma_s}, \frac{I_i}{\sigma_c})
    \label{eq:gray_filter}
\end{equation}

\vspace{-6mm}
\begin{equation}
    \vec{v_i} = (r_i,g_i,b_i, 1),  \;
    \vec{p_i} = (\frac{x_i}{\sigma_s}, \frac{y_i}{\sigma_s}, \frac{r_i}{\sigma_c}, \frac{g_i}{\sigma_c}, \frac{b_i}{\sigma_c})
    \label{eq:color_filter}
\end{equation}

\vspace{-2mm}
\noindent
where $exp^{-\frac{1}{2}(\vec{p_i} - \vec{p_j})^T\sum^{-1}(\vec{p_i}-\vec{p_j})^T }$ is a Gaussian distribution denoting the weight of the neighbor signal value $\vec{v_j}$ contributing to the target signal $\vec{v_i}$. Here $\vec{v}$ is the value vector of the signal, and $\vec{p}$ is the position vector of the signal. Equation \ref{eq:gaussian_blur}, \ref{eq:gray_filter} and \ref{eq:color_filter} denote Gaussian blur filter, gray-scale bilateral filter and color bilateral filter, respectively, where $v_i$ and $p_i$ are defined in Eq. \ref{eq:gaussian}. For these image filters, the signals are pixels, the signal values are 3D homogeneous color space, and signal positions are 2D, 3D and 5D, respectively, and the filtering processing is on 2D image Cartesian coordinate. The dimension of the position vector can be extended to $d$, which is called a \text{d-dimensional Gaussian filter}. We refer the readers to \cite{fast_filtering, BCL} and our supplementary materials for more details.

\noindent
\textbf{Permutohedral Lattice Network:}
However, a high-dimensional Gaussian filter can be also implemented on features with $n$ dimension rather than on pixels with 3D color space. The filtering process can be implemented in a more efficient space, the Permutohedral Lattice rather than a traditional Cartesian 2D image plane or 3D space. The $d$-dimensional permutohedral lattice is defined as the projection of the scaled Cartesian $(d+1)$-dimensional grid along the vector $\vec{1} = [1,1 ...1]$ onto the hyperplane $H_d$, which is the subspace of $R_{d+1}$ in which coordinates sum to zero:
\vspace{-1mm}
\begin{equation}
B_d = 
\begin{bmatrix}
d & -1 & ... & -1\\
-1 & d & ... & -1\\
... & ... & ... & ...\\
-1 & -1 & ... & d
\end{bmatrix}
\label{eq:bd}
\end{equation}
\vspace{-2mm}

Therefore it is also spanned by the projection of the standard basis for the original $(d+1)$-dimensional Cartesian space to $H_d$. In other words, we can project a feature at Cartesian coordinate $p_{cart}$ to Permutohedral Lattice coordinate $p_{lat}$ as:

\vspace{-7mm}
\begin{equation}
    \vec{p_{lat}} = {B_d} \vec{p_{cart}}
    \label{eq:3dtolattice}
\end{equation}
\vspace{-6mm}

With features embedding on lattice, \cite{fast_filtering} derived a maneuver pipeline ``splatting-convolution-slicing" for feature processing in lattice; \cite{BCL} improved the pipeline to be learnable BCL (Bilateral Convolutional Layers). Following CNN notion, \cite{hplflownet} improved BCL to different levels of receptive fields. 

To summarize, Permutohedral Lattice Network convolves the feature values $\vec{v_i}$ based on feature positions $\vec{p_i}$. While it is straightforward to derive $\vec{v_i}$ and $\vec{p_i}$ when feature positions and lattice have equal dimension, eg. 3D Lidar points input to 3D lattice\cite{BCL,splatnet,hplflownet} or 2D to 2D \cite{semantic_lattic, fast_filtering}, it is challenging when dimensions are different, as in the case of our 2D image input to 3D lattice. The proposed MonoPLFlowNet is designed to overcome such problem.

\subsection{Depth Decoder}

\begin{figure*}[t]
    \centering
    \includegraphics[width=\linewidth]{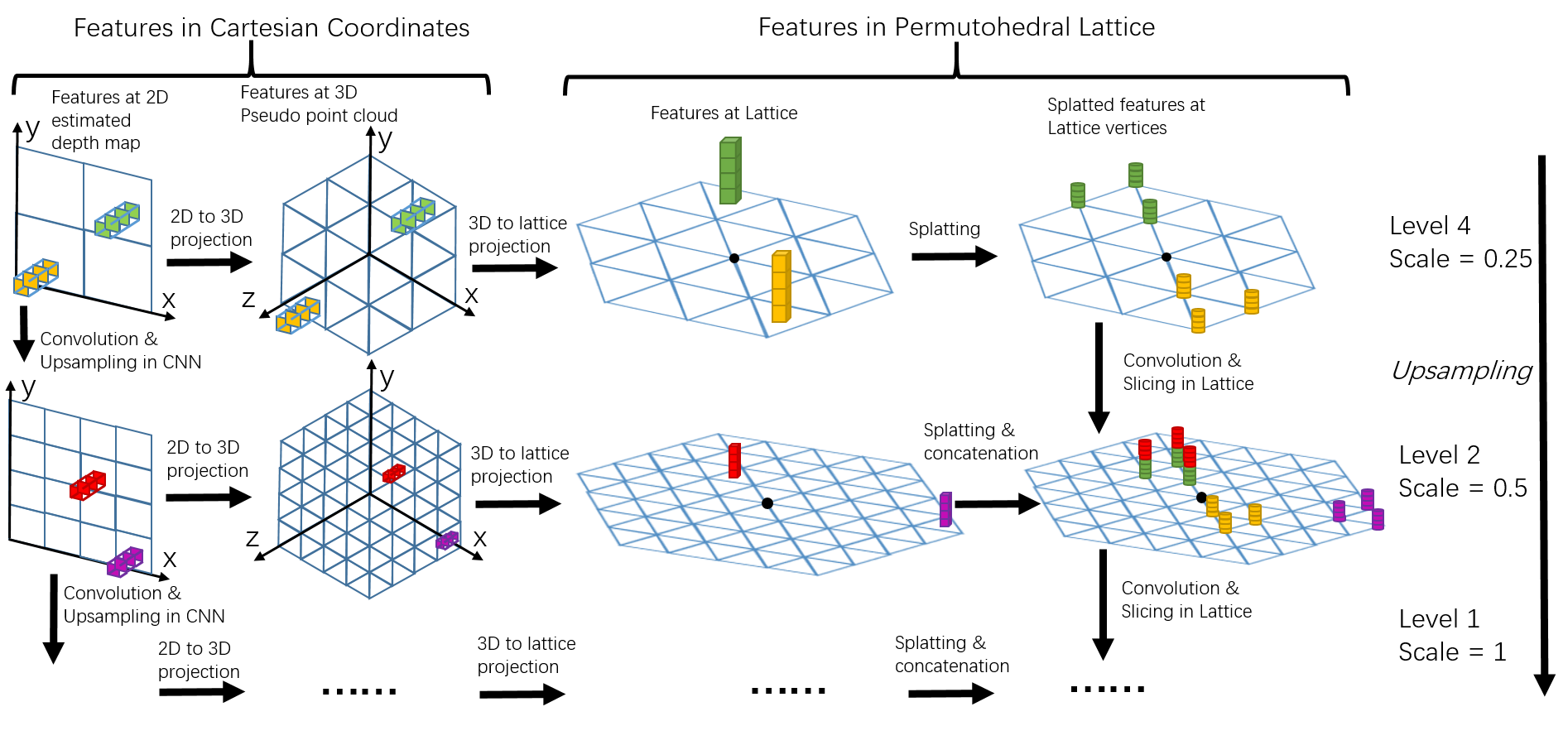}
    \vspace{-10mm}
    \caption{\textbf{3D Scene Flow Decoder: Pyramid-Level 2D-3D features alignment between Cartesian Coordinate and Permutohedral Lattice Network.} Horizontally it shows the feature embedding from Cartesian coordinate to permutohedral lattice. Vertically, it shows how the features are upsampled and concatenated for the next level.}
    \label{fig:sceneflow_decoder}
    \vspace{-5mm}
\end{figure*}

We design our depth module as an encoder-decoder based network as shown in Figure \ref{fig:depth_decoder}. Following the success of BTS DepthNet\cite{BTS}, we use their same encoder as a CNN-based feature extractor followed by the dilated convolution (ASPP)\cite{dense_aspp,aspp}. Our major contribution focuses on the decoder from three aspects.  

\noindent
\textbf{Level-Based Decoder:}
First, while keeping a pyramid-level top-down reasoning, BTS designed ``lpg" (local planar guidance) to regress depth at each level. In our experiments, we found that ``lpg" is a block where accuracy is sacrificed for efficiency. Instead, we replace the ``lpg" with our ``level-based decoder" as shown in Fig. \ref{fig:depth_decoder}, and improve the decoder to accommodate our sceneflow task.   

\noindent  
\textbf{Pyramid-Level Linearly-Additive mechanism:} BTS concatenates the estimated depth at each level and uses a final convolution to regress the final depth in a non-linear way, implying that the estimated depth at each level is not in a fixed scale. Instead, we propose a ``Pyramid-Level Linearly-Additive mechanism" as:

\vspace{-5mm}
\begin{equation}
    d_{final} = \frac{1}{4}(1 \times d_{lev1} + 2 \times d_{lev2} + 4 \times d_{lev4} + 8 \times d_{lev8})
    \label{eq:linear_additive}
\end{equation}
\vspace{-5mm}

\noindent
such that the final estimation is a linear combination over each level depth, where we force $d_{lev1}$, $d_{lev2}$, $d_{lev4}$, $d_{lev8}$ to be in $1, \frac{1}{2}, \frac{1}{4}, \frac{1}{8}$ scale of the real-scale depth. In order to achieve this, we also need to derive a pyramid-level loss corresponding to the level of the decoder architecture to supervise the depth at each level as Eq. \ref{eq:depth-loss}.  

Experiments show that with our improvement, our depth decoder outperforms the baseline BTS as well as other state-of-the-art works. More importantly, the objective is to design the depth decoder to generate feature/lattice for the scene flow decoder, which is our major contribution. More detail is provided in the next section.

\subsection{Scene Flow Decoder}
Our scene flow decoder is designed as a two-stream pyramid-level model which decodes in parallel in two domains - 2D Cartesian coordinate and 3D permutalhedral lattice, as shown in Figure \ref{fig:sceneflow_decoder}. As mentioned in Section 3.1, feature values and positions are two key factors in filtering processing. For the feature values, we use the features decoded by level-based decoder from depth module as shown in Figure \ref{fig:depth_decoder}, the features as the input to scene decoder (32 dimensions) are before regressing to the corresponding level depth (purple arrow), which means the input features to the scene decoder are the high-level feature representation of the corresponding level depth. While values are trivial, positions are not straightforward. As the challenge mentioned in Section 3.1, the 2D feature position from our depth is not enough to project the feature values to 3D permutohedral lattice. We derive a novel strategy to overcome this challenge by projecting the estimated depth estimation into 3D space to generate real-scale pseudo point cloud. Since pseudo point cloud has real-scale 3D coordinates, we can project the feature values from the 2D depth map to 3D point cloud in Cartesian coordinate, and then project to the corresponding position in 3D permutohedral lattice. A 2D to 3D projection is defined as:

% depth table on KITTI
% \usepackage{multirow}
\begin{table*}
\resizebox{\textwidth}{!}
{\begin{tabular}{r|c|c|ccc|cccc}
\hline
\multicolumn{1}{c|}{\multirow{2}{*}{Method}} & \multirow{2}{*}{Output} & \multirow{2}{*}{Scale} & \multicolumn{3}{c|}{\textit{higher is better}}   & \multicolumn{4}{c}{\textit{lower is better}}                      \\ \cline{4-10} 
\multicolumn{1}{c|}{}                        &                             &                             & $\delta <$ 1.25         & $\delta <$ 1.25         & $\delta <$ 1.25         & Abs Rel        & Sq Rel         & RMSE           & RMSE \textit{log}       \\ \hline
Make3D\cite{Make3D}                                       & D                           & \checkmark                           & 0.601          & 0.820          & 0.926          & 0.280          & 3.012          & 8.734          & 0.361          \\
Eigen et al.\cite{depth_eigen}                                 & D                           & \checkmark                            & 0.702          & 0.898          & 0.967          & 0.203          & 1.548          & 6.307          & 0.282          \\
Liu et al.\cite{depth_liu}                                   & D                           & \checkmark                            & 0.680          & 0.898          & 0.967          & 0.201          & 1.584          & 6.471          & 0.273          \\
LRC (CS + K)\cite{LRC}                                 & D                           & \checkmark                            & 0.861          & 0.949          & 0.976          & 0.114          & 0.898          & 4.935          & 0.206          \\
Kuznietsov et al.\cite{depth_kuznietsov}                            & D                           & \checkmark                            & 0.862          & 0.960          & 0.986          & 0.113          & 0.741          & 4.621          & 0.189          \\
Gan et al.\cite{depth_gan}                                   & D                           & \checkmark                            & 0.890          & 0.964          & 0.985          & 0.098          & 0.666          & 3.933          & 0.173          \\
DORN\cite{DORN}                                         & D                           & \checkmark                            & 0.932          & 0.984          & 0.994          & 0.072          & 0.307          & 2.727          & 0.120          \\
Yin et al.\cite{depth_yin}                                   & D                           & \checkmark                            & 0.938          & 0.990          & 0.998          & 0.072          & -              & 3.258          & 0.117          \\
BTS (DenseNet-121)\cite{BTS}                           & D                           & \checkmark                            & 0.951          & 0.993          & 0.998          & 0.063          & 0.256          & 2.850          & 0.100          \\
BTS (ResNext-101)\cite{BTS}                            & D                           & \checkmark                            & 0.956          & 0.993          & 0.998          & \textbf{0.059}          & 0.245          & 2.756          & 0.096          \\ \hline
GeoNet\cite{geonet}                                       & D + 2DF                       & \ding{55}                           & 0.793          & 0.931          & 0.973          & 0.155          & 1.296          & 5.857          & 0.233          \\
DFNet\cite{dfnet}                                        & D + 2DF                       & \ding{55}                            & 0.818          & 0.943          & 0.978          & 0.146          & 1.182          & 5.215          & 0.213          \\
CC\cite{CC}                                           & D + 2DF                       & \ding{55}                            & 0.826          & 0.941          & 0.975          & 0.140          & 1.070          & 5.326          & 0.217          \\
GLNet\cite{GLNet}                                        & D + 2DF                       & \ding{55}                            & 0.841          & 0.948          & 0.980          & 0.135          & 1.070          & 5.230          & 0.210          \\
EPC\cite{epc}                                          & D + 2DF                       & \ding{55}                            & 0.847          & 0.926          & 0.969          & 0.127          & 1.239          & 6.247          & 0.214          \\
EPC++\cite{epc++}                                        & D + 2DF                       & \ding{55}                            & 0.841          & 0.946          & 0.979          & 0.127          & 0.936          & 5.008          & 0.209          \\
Mono-SF\cite{monosf}                                      & D + 3DF                       & \ding{55}                            & 0.851          & 0.950          & 0.978          & 0.125          & 0.978          & 4.877          & 0.208          \\ \hline
\textbf{Ours (DenseNet-121)}                 & \textbf{D + 3DF}              & \checkmark                & \textbf{0.960} & \textbf{0.994} & \textbf{0.999} & 0.060 & \textbf{0.230} & \textbf{2.627} & \textbf{0.095} \\ \hline
\end{tabular}}
\vspace{-3mm}
\caption{\textbf{Monocular depth results comparison on KITTI Eigen's split.} In the column \textit{Output}, \textit{D} denotes depth, \textit{2DF} and \textit{3DF} denote 2D optical flow and 3D scene flow. In the column \textit{Scale}, \checkmark denotes in real scale, \ding{55} denotes with scale ambiguity. DenseNet-121 as backbone for efficiency.}
\label{table:depth on kitti}
\end{table*}

\vspace{-2mm}
\begin{equation}
    \begin{bmatrix}
x\\
y\\
z
    \end{bmatrix}
    = 
    \begin{bmatrix}
{z}/{f_u} & 0 & {-c_u}z/{f_u}\\
0 & {z}/{f_v} & {-c_v}z/{f_v}\\
0 & 0 & z
    \end{bmatrix}
    \begin{bmatrix}
    u \\
    v \\
    1
    \end{bmatrix}
\label{eq:2d23d}
\end{equation}
\vspace{-2mm}

\noindent
where $z$ is the estimated depth, $(u, v)$ is the corresponding pixel coordinate in the depth map, $f_u, f_v$ are horizontal and vertical camera focal lengths, $(c_u, c_v)$ are the coordinate of camera principle point. $(x, y, z)$ is the coordinate of the projected 3D point. 

So far we have successfully derived the projection in the real scale, where the complete projection pipeline consisting of 2D to 3D (Eq. \ref{eq:2d23d}) and 3D to lattice (Eq. \ref{eq:3dtolattice}) is shown in Figure 4.  Due to the linear property of the projection, the proposed projection pipeline holds at different scales in the overall system. Using this property, we prove a stronger conclusion: scaling the feature depth from Cartesian coordinate leads to a same scale to the corresponding feature position in permutohedral lattice. Eq. \ref{eq:scene} summarizes the mapping where $(p_x, p_y, p_z)$ is the permutohedral lattice coordinate of the feature corresponding to its 2D Cartisian coordinate in the depth map. 

\vspace{-4mm}
\begin{equation}
\begin{split}
    \begin{bmatrix}
p^\prime_x\\
p^\prime_y\\
p^\prime_z
    \end{bmatrix}
    = B_d
    \begin{bmatrix}
    x^\prime \\
    y^\prime \\
    z^\prime
    \end{bmatrix} \\
    = B_d
    \begin{bmatrix}
{z^\prime}/{f^\prime_u} & 0 & {-c^\prime_u}z^\prime/{f^\prime_u} \\
0 & {z^\prime}/{f^\prime_v} & {-c^\prime_u}z^\prime/{f^\prime_u} \\
0 & 0 & z^\prime
    \end{bmatrix}
    \begin{bmatrix}
u^\prime \\
v^\prime \\
1
    \end{bmatrix} \\
    = B_d
    \begin{bmatrix}
{\lambda z}/{\lambda f_u} & 0 & {-\lambda c_u}\lambda z/{\lambda f_u} \\
0 & {\lambda z}/{\lambda f_v} & {-\lambda c_u}\lambda z/{\lambda f_u} \\
0 & 0 & \lambda z
    \end{bmatrix}
    \begin{bmatrix}
u^\prime \\
v^\prime \\
1
    \end{bmatrix} \\
    = B_d
    \begin{bmatrix}
{z}/{f_u} & 0 & {-c_u}\lambda z/{f_u} \\
0 & {z}/{f_v} & {-c_u}\lambda z/{f_u} \\
0 & 0 & \lambda z
    \end{bmatrix}
    \begin{bmatrix}
\lambda u \\
\lambda v \\
1
    \end{bmatrix} \\
    = B_d
    \begin{bmatrix}
{\lambda z}/{f_u} & 0 & {-c_u}\lambda z/{f_u} \\
0 & {\lambda z}/{f_v} & {-c_u}\lambda z/{f_u} \\
0 & 0 & \lambda z
    \end{bmatrix}
    \begin{bmatrix}
u\\
v\\
1
    \end{bmatrix} \\
\end{split}
\label{eq:scene}
\end{equation}
\vspace{-3mm}

\noindent
$B_d$ is defined in Eq. \ref{eq:bd}, $\lambda$ is the scale, a ``prime" sign denotes scaling the original value with $\lambda$. Comparing the final result of Eq. \ref{eq:scene} to Eq. \ref{eq:2d23d}, the only difference is to replace $z$ with $\lambda z$, hence only scaling the depth in Cartesian coordinate will lead to a same scale to the position in permutohedral lattice. Using the proposed ``Pyramid-Level 2D-3D features alignment" mechanism, we can embed the features from 2D Cartesian coordinate to 3D Permutohetral lattice, and then implement splating-convolution-slicing and concatenate different level features directly in the permutohedral lattice network. Please see the supplementary materials for more explanation. For the basic operations in permutohedral lattice, we directly refer to \cite{hplflownet}.

\subsection{Loss}

%depth fly
\begin{table*}[]
\begin{tabular}{r|c|c|ccc|cccc}
\hline
\multicolumn{1}{c|}{\multirow{2}{*}{Method}} & \multirow{2}{*}{Train on} & \multirow{2}{*}{Scale} & \multicolumn{3}{c|}{\textit{higher is better}} & \multicolumn{4}{c}{\textit{lower is better}} \\ \cline{4-10} 
\multicolumn{1}{c|}{}                        &                           &                             & $\delta <$ 1.25           & $\delta <$ 1.25          & $\delta <$ 1.25          & Abs Rel   & Sq Rel   & RMSE     & RMSE log   \\ \hline
Mono-SF\cite{monosf}                                      & K                     & \ding{55}                       & 0.259          & 0.483         & 0.648         & 0.943     & 19.250   & 14.676   & 0.667      \\
Mono-SF-Multi\cite{monosf_multi}                                & K                     & \ding{55}                        & 0.273          & 0.492         & 0.650         & 0.931     & 19.072   & 14.566   & 0.666      \\
\textbf{Ours}                                         & F           & \checkmark                        & \textbf{0.715}          & \textbf{0.934}         & \textbf{0.980}         & \textbf{0.188}     & \textbf{1.142}    & \textbf{4.400}    & \textbf{0.235}      \\ \hline
\end{tabular}
\vspace{-3mm}
\caption{\textbf{Monocular depth results comparison on Flyingthings3D dataset.} In the column \textit{Train on}, K denotes KITTI, F denotes Flyingthings3D. In the column \textit{Scale}, \checkmark denotes in real scale, \ding{55} denotes with scale ambiguity.}
\label{table:depth on fly}
\end{table*}

%scene fly
\begin{table*}[]
\begin{tabular}{ccccccccc}
\hline
\multicolumn{1}{c|}{Method}        & \multicolumn{1}{c|}{Train on}       & \multicolumn{1}{c|}{Scale} & EPE3D(m)             & ACC3DS               & ACC3DR               & \multicolumn{1}{c|}{Outlier3D} & EPE2D(px)            & ACC2D                \\ \hline
\multicolumn{1}{c|}{Mono-SF\cite{monosf}}       & \multicolumn{1}{c|}{K}          & \multicolumn{1}{c|}{\ding{55}}       & 1.1288               & 0.0525               & 0.1017               & \multicolumn{1}{c|}{0.9988}    & 58.2761              & 0.2362               \\
\multicolumn{1}{c|}{Mono-SF-Multi \cite{monosf_multi}} & \multicolumn{1}{c|}{K}          & \multicolumn{1}{c|}{\ding{55}}       & 1.5864               & 0.0020               & 0.0050               & \multicolumn{1}{c|}{0.9988}    & 48.3099              & 0.3162               \\ \hline
\multicolumn{1}{c|}{\textbf{Ours}}          & \multicolumn{1}{c|}{F} & \multicolumn{1}{c|}{\checkmark}       & \textbf{0.3915}               & \textbf{0.5424}               & \textbf{0.6911}               & \multicolumn{1}{c|}{\textbf{0.8279}}    & \textbf{22.4226}              & \textbf{0.6659}               \\ \hline
\multicolumn{1}{l}{}               & \multicolumn{1}{l}{}                & \multicolumn{1}{l}{}            & \multicolumn{1}{l}{} & \multicolumn{1}{l}{} & \multicolumn{1}{l}{} & \multicolumn{1}{l}{}           & \multicolumn{1}{l}{} & \multicolumn{1}{l}{} \\
\multicolumn{1}{l}{}               & \multicolumn{1}{l}{}                & \multicolumn{1}{l}{}            & \multicolumn{1}{l}{} & \multicolumn{1}{l}{} & \multicolumn{1}{l}{} & \multicolumn{1}{l}{}           & \multicolumn{1}{l}{} & \multicolumn{1}{l}{}
\end{tabular}
\vspace{-12mm}
\caption{\textbf{Monocular 3D scene flow results comparison on Flyingthings3D dataset.(image based evaluation standard)} In the column \textit{Train on}, K denotes KITTI, F denotes Flyingthings3D. In the column \textit{Scale}, \checkmark denotes in real scale, \ding{55} denotes with scale ambiguity.}
\label{table:scene on fly}
\end{table*}

%scene kitti
\begin{table*}
\resizebox{\textwidth}{!}
{\begin{tabular}{ccccccccc}
\hline
\multicolumn{1}{c|}{Method}                         & \multicolumn{1}{c|}{Train on}       & \multicolumn{1}{c|}{Scale} & EPE3D(m)             & ACC3DS               & ACC3DR               & \multicolumn{1}{c|}{Outlier3D} & EPE2D(px)            & ACC2D                \\ \hline
\multicolumn{1}{c|}{\cite{monosf}depth + Mono-Exp\cite{mono_expansion}} & \multicolumn{1}{c|}{K}          & \multicolumn{1}{c|}{\ding{55}}  & 2.7079               & 0.0676               & 0.1467               & \multicolumn{1}{c|}{0.9982}    & 181.0699             & 0.2777               \\
\multicolumn{1}{c|}{Our depth + Mono-Exp\cite{mono_expansion}}    & \multicolumn{1}{c|}{K}          & \multicolumn{1}{c|}{\ding{55}}  & 1.6673               & 0.0838               & 0.1815               & \multicolumn{1}{c|}{0.9953}    & 78.7245              & 0.2837               \\
\multicolumn{1}{c|}{Mono-SF\cite{monosf}}                        & \multicolumn{1}{c|}{K}          & \multicolumn{1}{c|}{\ding{55}}  & 1.1288               & 0.0525               & 0.1017               & \multicolumn{1}{c|}{0.9988}    & 58.2761              & 0.2362               \\
\multicolumn{1}{c|}{Mono-SF-Multi\cite{monosf_multi}}                  & \multicolumn{1}{c|}{K}          & \multicolumn{1}{c|}{\ding{55}}  & 0.7828               & 0.1725               & 0.2548               & \multicolumn{1}{c|}{0.9477}    & 35.9015              & 0.4886               \\ \hline
\multicolumn{1}{c|}{\textbf{Ours}}                           & \multicolumn{1}{c|}{F} & \multicolumn{1}{c|}{\checkmark}  & \textbf{0.6970}               & \textbf{0.2453}               & \textbf{0.3692}               & \multicolumn{1}{c|}{\textbf{0.8630}}    & \textbf{33.4750}              & \textbf{0.4968}               \\ \hline
\multicolumn{1}{l}{}                                & \multicolumn{1}{l}{}                & \multicolumn{1}{l}{}       & \multicolumn{1}{l}{} & \multicolumn{1}{l}{} & \multicolumn{1}{l}{} & \multicolumn{1}{l}{}           & \multicolumn{1}{l}{} & \multicolumn{1}{l}{} \\
\multicolumn{1}{l}{}                                & \multicolumn{1}{l}{}                & \multicolumn{1}{l}{}       & \multicolumn{1}{l}{} & \multicolumn{1}{l}{} & \multicolumn{1}{l}{} & \multicolumn{1}{l}{}           & \multicolumn{1}{l}{} & \multicolumn{1}{l}{}
\end{tabular}}
\vspace{-12mm}
\caption{\textbf{Monocular 3D scene flow results comparison on KITTI flow 2015 dataset.(image based evaluation standard)} In the column \textit{Train on}, K denotes KITTI, F denotes Flyingthings3D. In the column \textit{Scale}, \checkmark denotes in real scale, \ding{55} denotes with scale ambiguity.}
\label{table:scene on kitti}
\end{table*}

% scene ablation
\begin{table*}[]
\begin{tabular}{cclccccc}
\hline
\multicolumn{1}{c|}{Method}                  & EPE3D(m)             & Scale & ACC3DS               & ACC3DR               & \multicolumn{1}{c|}{Outlier3D} & EPE2D(px)            & ACC2D                \\ \hline
\multicolumn{1}{c|}{MonoPLFlowNet-lev1}      & 0.4781               & \checkmark  & 0.4587               & 0.6146               & \multicolumn{1}{c|}{0.8935}    & 26.3133              & 0.6092               \\
\multicolumn{1}{c|}{MonoPLFlowNet-lev1-lev2} & 0.4439               & \checkmark  & 0.4689               & 0.6333               & \multicolumn{1}{c|}{0.8605}    & 24.3198              & 0.6366               \\
\multicolumn{1}{c|}{\textbf{MonoPLFlowNet-full}}      & \textbf{0.4248}               & \checkmark  & \textbf{0.5099}               & \textbf{0.6611}               & \multicolumn{1}{c|}{\textbf{0.8595}}    & \textbf{23.7657}              & \textbf{0.6456}               \\ \hline
\multicolumn{1}{l}{}                         & \multicolumn{1}{l}{} &       & \multicolumn{1}{l}{} & \multicolumn{1}{l}{} & \multicolumn{1}{l}{}           & \multicolumn{1}{l}{} & \multicolumn{1}{l}{} \\
\multicolumn{1}{l}{}                         & \multicolumn{1}{l}{} &       & \multicolumn{1}{l}{} & \multicolumn{1}{l}{} & \multicolumn{1}{l}{}           & \multicolumn{1}{l}{} & \multicolumn{1}{l}{}
\end{tabular}
\vspace{-11mm}
\caption{\textbf{Ablation study on our MonoPLFlowNet by changing level of the scene decoder.(image based evaluation standard)} lev1 denotes only using the last level, lev1-lev2 denotes using the last two levels, full denotes using all levels. For fair comparison, we show all results after training epoch 22.}
\label{table:scene ablation}
\textbf{\vspace{-5mm}}
\end{table*}

\noindent
\textbf{Depth Loss:}
\noindent
Silog (scale-invariant log) loss is a widely-used loss\cite{depth_eigen} for depth estimation supervision defined as:
\vspace{-2mm}
\begin{equation}
    L_{silog}(\Tilde{d}, d) = \alpha \sqrt{\frac{1}{T}\sum_i(g_i)^2 - \frac{\lambda}{T^2}(\sum_i g_i)^2}
    \vspace{-2mm}
\end{equation}
where $g_i = \log \Tilde{d_i}-\log d_i$, $\Tilde{d}$ and $d$ are estimated and ground truth depths, $T$ is the number of valid pixels, $\lambda$ and $\alpha$ are constants set to be 0.85 and 10. Since our depth decoder decodes a fixed-scale depth at each level, we do not directly supervise on final depth. Instead, we design a pyramid-level silog loss corresponding to our depth decoder to supervise the estimation from each level. 
\vspace{-2mm}
\begin{equation}
    L_{depth} = \frac{1}{15}(8\times L_{1} + 4\times L_{2} + 2\times L_{4} + 1\times L_{8})
    \vspace{-2mm}
    \label{eq:depth-loss}
\end{equation}
where $L_{level} = L_{silog}(\Tilde{d}_{level}, d_{level}/n)$. Higher weight is assigned to low-level loss to stabilize the training process.
\vspace{2mm}  

\noindent
\textbf{Scene Flow Loss:}
Following most LiDAR-based work, we first use a traditional End Point Error (EPE3D) loss as $L_{epe} = || \Tilde{sf} - sf ||_2$, where $\Tilde{sf}$ and $sf$ are estimated and ground truth scene flows, respectively. To bring two sets of point clouds together, some self-supervised works leverage the Chamfer distance loss as:
\begin{equation}
    L_{cham}(P, Q) = \sum_{p\in P} \min_{q\in Q}||p-q||_2^2 + \sum_{q\in Q} \min_{p\in P}||p-q||_2^2
\end{equation}

\noindent
where $P$ and $Q$ are two sets of point clouds that optimized to be close to each other. While EPE loss supervises directly on scene flow 3D vectors, we improved canonical Chamfer loss to a forward-backward Chamfer distance loss supervising on our pseudo point cloud from depth estimates as:
\vspace{-3mm}
\begin{equation}
\begin{split}
    L_{cham\textunderscore total} = L_{cham\textunderscore f} +  L_{cham\textunderscore b}\\
    L_{cham\textunderscore f} = L_{cham}(\Tilde{P}_{f}, P_2)\\
    L_{cham\textunderscore b} = L_{cham}(\Tilde{P}_{b}, P_1)\\
    \Tilde{P}_{f} = P_1 + \Tilde{sf}_{f}, \ \Tilde{P}_{b} = P_2 + \Tilde{sf}_{b}
\end{split}
\end{equation}
where $P_1$ and $P_2$ are pseudo point clouds generated from the estimated depth of two consecutive frames. $\Tilde{sf}_{f}$ and $\Tilde{sf}_{b}$ are estimated forward and backward scene flows. 
%-----------------------------------------------------------------------
\section{Experiments}

\noindent
\textbf{Datasets:} We use Flyingthings3D\cite{flyingthings3d} and KITTI\cite{kitti} dataset in this work for training and evaluation. Flyingthings3D is a synthetic dataset with 19460 pairs of images in its training split, and 3824 pairs of images in its evaluation split. We use it for training and evaluation of both depth and scene flow estimation. For KITTI dataset, following most previous works, for depth training and evaluation, we use KITTI Eigen's\cite{depth_eigen} split which has 23488 images of 32 scenes for training and 697 images of 29 scenes for evaluation. For scene flow evaluation, we use KITTI flow 2015\cite{kittiflow} split with 200 pairs of images labeled with flow ground truth. We do not train scene flow on KITTI. 

\subsection{Monocular Depth}
We train the depth module in a fully-supervised manner using the pyramid-level silog loss derived in Eq. \ref{eq:depth-loss}. For training simplicity, we first train the depth module free from scene flow decoder. Our model is completely trained from scratch.

\textbf{KITTI:} We first train on KITTI Eigen's training split, Table \ref{table:depth on kitti} shows the depth comparison on KITTI Eigen's evaluation split. We classify the previous works into two categories, joint-estimate monocular depth and flow, and single-estimate monocular depth alone. It is clearly from the table that the single estimation outperforms the joint estimation on average, and the joint estimation has scale ambiguity. With a similar design to our strongest single-estimate baseline BTS\cite{BTS}, our depth outperforms BTS with the same backbone DenseNet-121\cite{densenet} as well as the best BTS with ResNext-101 backbone. The joint-estimate works fail to estimate in real scale because they regress depth and scene flow together in self-supervised manner, which sacrifice the real depth. We design our model with separate decoders in a fully-supervised manner, and succeed to jointly estimate depth and scene flow in real scale, where our depth achieves a big improvement to the strongest joint-estimate baseline Mono-SF\cite{monosf}. The monocular depth evaluation metrics cannot show the difference of a normalized and real-scale depth, but real-scale depth is required for real-scale 3D scene flow estimation.

\textbf{Flyingthings3D:} We use the same way to train another version on Flyingthings3D from scratch, because we need to use this version to train the scene flow decoder on Flyingthings3D. Since Flyingthings3D is not a typical dataset for depth training, very few previous works reported results. Because scene flow estimation is related to depth, we also evaluated two strongest baselines on Flyingthings3D as shown in Table \ref{table:depth on fly}. Then we use the trained depth module to train the scene flow decoder. 

\subsection{Real-Scale 3D Scene Flow}

% scene all
\begin{table*}[]
\begin{tabular}{r|r|c|cccc|cc}
\hline
\multicolumn{1}{c|}{Dataset} & \multicolumn{1}{c|}{Method} & Input         & EPE3D(m) & ACC3DS & ACC3DR & Outlier3D & EPE2D(px) & ACC2D  \\ \hline
                             & LDOF\cite{LDOF}                        & RGBD          & 0.498    & -      & -      & -         & -         & -      \\
                             & OSF\cite{OSF}                         & RGBD          & 0.394    & -      & -      & -         & -         & -      \\
                             & PRSM\cite{PRSM}                        & RGBD          & 0.327    & -      & -      & -         & -         & -      \\
                             & PRSM\cite{PRSM}                        & Stereo   & 0.729    & -      & -      & -         & -         & -      \\
K                   
                             & Ours                        & Mono &  0.6970        &   0.0035    &   0.0255    &  0.9907         &  33.4750         &  0.0330      \\
                             & ICP(rigid)\cite{icp}                  & LiDAR  & 0.5185   & 0.0669 & 0.1667 & 0.8712    & 27.6752   & 0.1056 \\
                             & FGR(rigid)\cite{fgr}                  & LiDAR  & 0.4835   & 0.1331 & 0.2851 & 0.7761    & 18.7464   & 0.2876 \\
                             & CPD(non-rigid)\cite{cpd}              & LiDAR  & 0.4144   & 0.2058 & 0.4001 & 0.7146    & 27.0583   & 0.1980 \\ \hline
                             & FlowNet-C\cite{flownet-c}                   & Depth & 0.7887   & 0.0020 & 0.0149 & -         & -         & -      \\
                             & FlowNet-C\cite{flownet-c}                   & RGBD          & 0.7836   & 0.0025 & 0.0174 & -         & -         & -      \\
                             & Ours                        & Mono         & 0.3915   & 0.0125 & 0.0816 & 0.9874    & 22.4226   & 0.0724 \\
               & ICP(global)\cite{icp}                  & LiDAR & 0.5019   & 0.0762 & 0.2198 & -         & -         & -      \\
               F
                             & ICP(rigid)\cite{icp}                  & LiDAR & 0.4062   & 0.1614 & 0.3038 & 0.8796    & 23.2280   & 0.2913 \\
                             & FlowNet3D-EM\cite{flownet3d}                & LiDAR  & 0.5807   & 0.0264 & 0.1221 & -         & -         & -      \\
                             & FlowNet3D-LM\cite{flownet3d}                & LiDAR & 0.7876   & 0.0027 & 0.0183 & -         & -         & -      \\
                             & FlowNet3D\cite{flownet3d}                   & LiDAR & 0.1136   & 0.4125 & 0.7706 & 0.6016    & 5.9740    & 0.5692 \\ \hline
\end{tabular}
\vspace{-3mm}
\caption{\textbf{3D scene flow results comparison with different input data form on KITTI flow 2015 and Flyingthings3D. (LiDAR based evaluation standard)} Since we also compare the LiDAR approaches here, we use the strict LiDAR-based evaluation standard. In the column \textit{Dataset}, K denotes KITTI, F denotes Flyingthings3D. All works in the table are in real scale. Since our work is the first image-based work thoroughly evaluating 3D scene flow with 3D metrics, we lack of some 3D results from previous image-based works, but it is already enough to see our Monocular-image based work is comparable to LiDAR approaches}
\label{table:scene all}
\end{table*}

\begin{figure*}
\begin{minipage}{0.45\textwidth}
  \centerline{\includegraphics[width=1.0\textwidth]{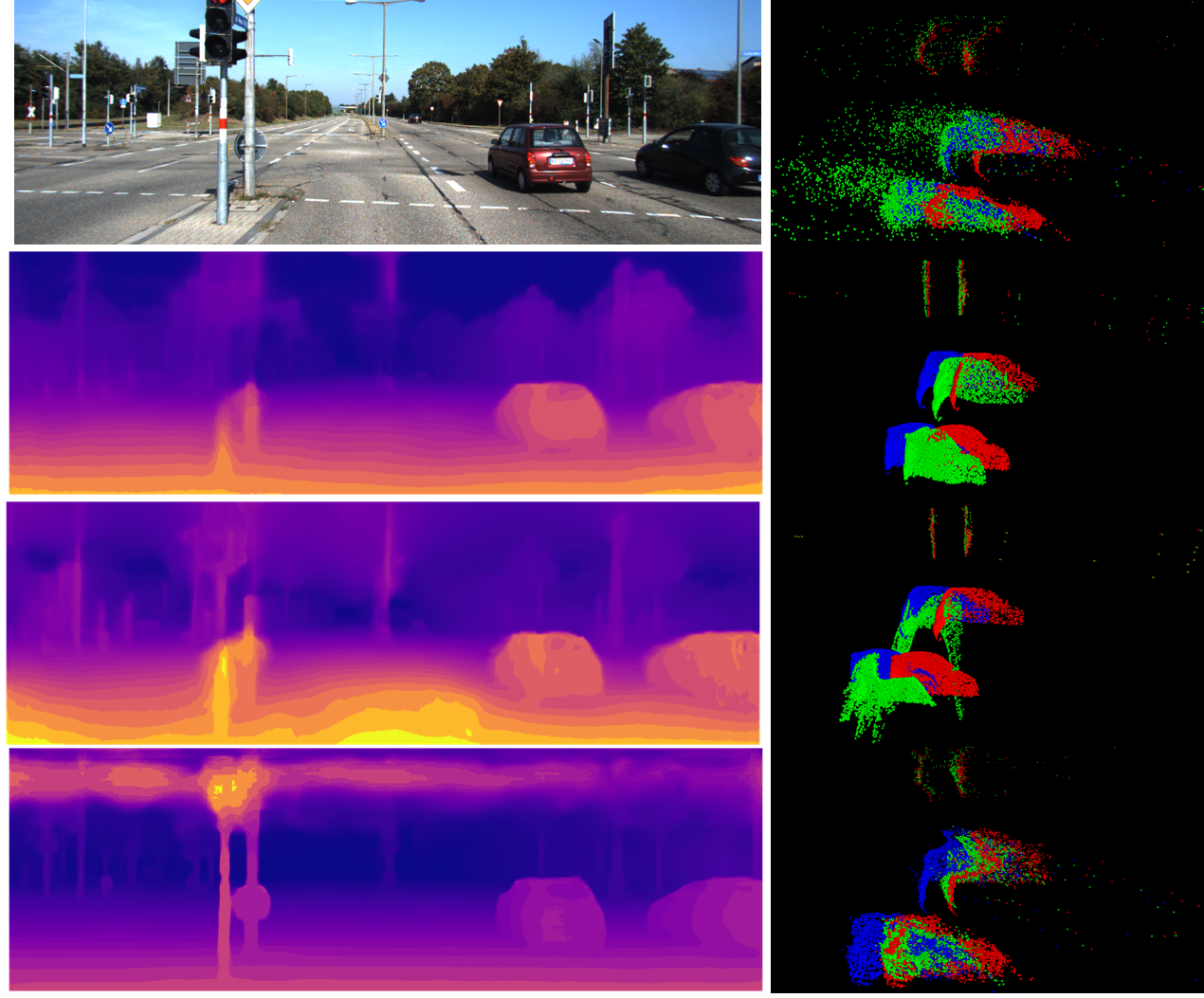}}
\end{minipage}
\hfill
\begin{minipage}{0.54\linewidth}
  \centerline{\includegraphics[width=1.0\textwidth]{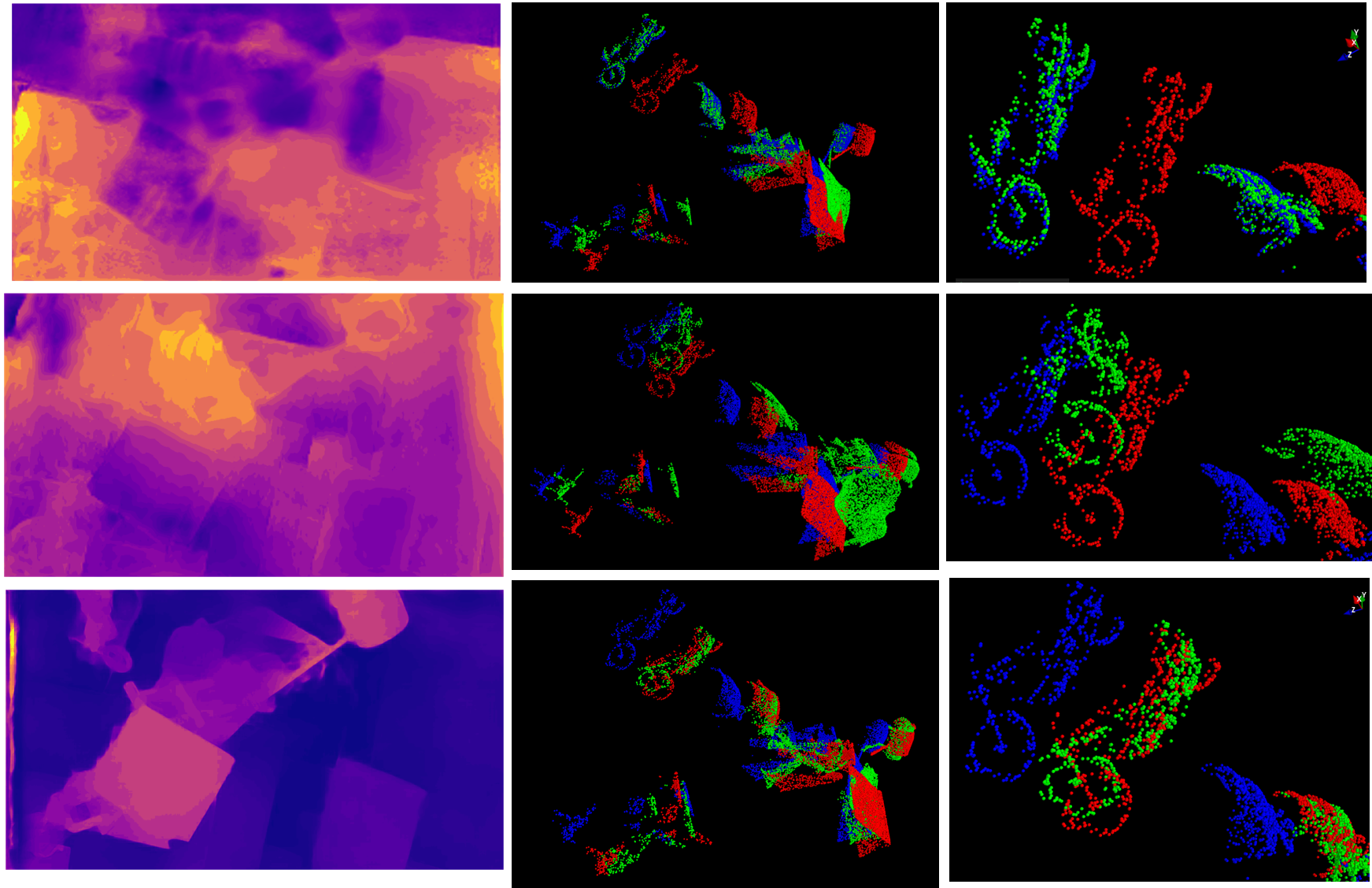}}
\end{minipage}
\vspace{-3mm}
\caption{\textbf{Qualitative depth and real-scale 3D scene flow results of the proposed MonoPLFlowNet on KITTI and Flyingthins3D for a single pair of two consecutive frames.} For KITTI (column 1\&2), 1st row: 1st frame of the RGB input image, recovered scene flow of \cite{mono_expansion} by our depth. From 2nd to 4th row: depth and scene flow by Mono-sf\cite{monosf}, Mono-sf-multi\cite{monosf_multi} and ours. For Flyingthings3D (column3\&4\&5), from top to down: depth of the 1st frame, scene flow, zoom-in view scene flow by \cite{monosf, monosf_multi} and ours, original input RGB is shown in supplementary materials. Depth and scene flow of \cite{monosf, monosf_multi} are recovered to the real scale before generating point cloud.}
\label{fig:visualization}
\vspace{-5mm}
\end{figure*}

Most image-based scene flow works are trained on KITTI in a self-supervised manner, which leads to the scale ambiguity. We train our scene flow decoder in a fully-supervised manner with the EPE3D and forward-backward loss proposed in Section 3.4, which is able to estimate real-scale depth and scene flow. While real dataset like KITTI lacks 3D scene flow label, we only train our depth decoder on synthetic dataset Flyingthings3D. 

Previous image-based scene flow works mostly use $d1, d2, f1, sf1$ for evaluation, but these metrics are designed for evaluating 2D optical flow or normalized scene flow, which themselves have the scale ambiguity. Instead, we use the metrics directly for evaluating real-scale 3D scene flow \cite{flownet3d, hplflownet, pointpwc}, which we refer as \textbf{LiDAR-based evaluation standard} (details in supplementary materials). However, since this LiDAR standard is too strict to image-based approaches, we slightly relax the LiDAR standard and use an \textbf{image-based evaluation standard}, EPE (end point error) 3D/2D are same to LiDAR standard:
\vspace{-2mm}
\begin{itemize}[leftmargin=0cm]
\setlength{\itemsep}{0pt}
\setlength{\parsep}{0pt}
\setlength{\parskip}{0pt}
\item Acc3DS: the percentage of points with EPE3D $<$ 0.3m or relative error $<$ 0.1.
\item Acc3DR: the percentage of points with EPE3D $<$ 0.4m or relative error $<$ 0.2.
\item Outliers3D: the percentage of points with EPE3D $>$ 0.5m or relative error
$>$ 0.3.
\item Acc2D: the percentage of points whose EPE2D $<$ 20px or relative error $<$ 0.2.
\end{itemize}
\vspace{-2mm}

Since we are the only monocular image-based approach estimating in real scale, to evaluate other works with our monocular approach, we need to recover other works to the real scale using the depth and scene flow ground truth (details in supplementary materials).

\textbf{Flyingthings3D:} Table \ref{table:scene on kitti} shows the monocular 3D scene flow comparison on Flyingthings3D. Even recovering \cite{monosf, monosf_multi} to real scale with ground truth, our result still outperforms the two strongest state-of-the-art baseline works by an overwhelming advantage, more important we only need two consecutive images without any ground truth.

\textbf{KITTI:} We also directly evaluate scene flow on KITTI without any  fine-tuning as shown in Table \ref{table:scene on kitti}. The table also includes state-of-the-art 2D approach Mono-Expansion\cite{mono_expansion}. We first recover its direct output 2D optical flow to 3D scene flow, and then recover the scale. To recover 2D to 3D, Mono-Expansion proposed a strategy using LiDAR ground truth to expand 2D to 3D specifically for its own usage, but it is not able to extend to all works. For comparison, we recover 3D flow and scale in the same way with ground truth. In the table, the proposed approach still outperforms all state-of-the-art strong baseline works without any fine-tuning on KITTI. Since Mono-Expansion does not estimate depth, we use our depth and Mono-SF depth to help recovering 3D scene flow. Note that our scene flow decoder does not use the depth directly, but share the features and regress in parallel. In the table, by using our depth to recover Mono-Exp, it greatly outperforms by using depth from Mono-SF \cite{monosf}. This comparison also shows the superiority of our depth estimation over others.
\vspace{-1mm}

\textbf{Ablation Study:} We perform the ablation study for our scene flow decoder. The ablation study verifies the 2D-3D features alignment process, discussed in Section 3.3. As our MonoPLFlowNet architecture (Figure \ref{fig:MonoPLFlowNet}) shows, we perform three-level 2D-3D features alignment in our full model and decode in parallel, which are level 1, 2, 4. In the ablation study, we train the model only with level1 and level1+level2, and compare to the full model trained to the same epoch. The results indicate that the performances get better with deeper levels involved, hence the concatenation of features from different levels in the lattice boost the training, which proves our 2D-3D features alignment mechanism.

\subsection{Visual Results}
We show our visual results of depth and real-scale scene flow in Figure \ref{fig:visualization}. 3D scene flow are visualized with the pseudo point cloud generating from the estimated depth map, where blue points are from 1st frame, red and green points are blue points translated to 2nd frame by ground truth and estimated 3D scene flow, respectively. The goal of the algorithm is to match the green points to the red points. Different to LiDAR-based works that have same shape of point cloud, the shapes of point cloud are different here because generating from different depth estimation. More visual results are in supplementary materials. 

%---------------------------------------------------------------
\section{Conclusion}
\vspace{-3.5mm}
We present MonoPLFlowNet in this paper.  It is the first deep learning architecture that can estimate both depth and 3D scene flow in real scale, using only two consecutive monocular images. Our depth and 3D scene flow estimation outperforms all the-state-of-art baseline monocular based works, and is comparable to LiDAR based works. In the future, we will explore the usage of more real datasets with specifically designed self-supervised loss to further improve the performance.

%%%%%%%%% REFERENCES
{\small
\bibliographystyle{ieee_fullname}
\bibliography{draft}
}

%-------------------------------------------------------------
\end{document}